\documentclass{article} 
\usepackage{iclr2026_conference,times}

\usepackage{amsmath,amsfonts,bm}

\def\eqref#1{equation~\ref{#1}}

\def\1{\bm{1}}

\DeclareMathAlphabet{\mathsfit}{\encodingdefault}{\sfdefault}{m}{sl}
\SetMathAlphabet{\mathsfit}{bold}{\encodingdefault}{\sfdefault}{bx}{n}

\usepackage{hyperref}
\usepackage{url}
\usepackage{booktabs}
\usepackage{graphicx}
\usepackage{caption}
\usepackage{subcaption}
\usepackage{multirow}

\usepackage{tabularx}

\title{Empirical Stability Analysis of Kolmogorov-Arnold Networks in Hard-Constrained Recurrent Physics-Informed Discovery}

\author{Enzo Nicol\'as Spotorno \\
Department of Informatics and Statistics \\
Federal University of Santa Catarina \\
\texttt{enzoniko@lisha.ufsc.br}
\AND
Josafat Leal Filho \\
Department of Informatics and Statistics \\
Federal University of Santa Catarina \\
\texttt{josafat@lisha.ufsc.br}
\AND
Ant\^onio Augusto Fr\"ohlich \\
Department of Informatics and Statistics \\
Federal University of Santa Catarina\\
\texttt{guto@lisha.ufsc.br}
}

\iclrfinalcopy 
\begin{document}

\maketitle

\begin{abstract}
We investigate the integration of Kolmogorov-Arnold Networks (KANs) into hard-constrained recurrent physics-informed architectures (HRPINN) to evaluate the fidelity of learned residual manifolds in oscillatory systems. Motivated by the Kolmogorov-Arnold representation theorem and preliminary gray-box results, we hypothesized that KANs would enable efficient recovery of unknown terms compared to MLPs. Through initial sensitivity analysis on configuration sensitivity, parameter scale, and training paradigm, we found that while small KANs are competitive on univariate polynomial residuals (Duffing), they exhibit severe hyperparameter fragility, instability in deeper configurations, and consistent failure on multiplicative terms (Van der Pol), generally outperformed by standard MLPs. These empirical challenges highlight limitations of the additive inductive bias in the original KAN formulation for state coupling and provide preliminary empirical evidence of inductive bias limitations for future hybrid modeling.
\end{abstract}


\section{Introduction}

Hard-constrained recurrent architectures, such as the Hybrid Recurrent Physics-Informed Neural Network (HRPINN)~\citep{spotorno2025hard}, embed known physics structurally within a recurrent integrator, forcing the network to learn \emph{only} residual dynamics. Specifically, the known structure and integrator are fixed in the recurrent update rule, while the neural residual branch learns exclusively the unknown manifold. This design ensures physical consistency by construction and was demonstrated to be effective for accuracy and invariant enforcement in cyber-physical systems. A key open direction identified in that work is the symbolic discovery of unknown residual terms.

Recently, Kolmogorov-Arnold Networks (KANs) \citep{liu2024kan} have emerged as a promising alternative for scientific machine learning. Grounded in the Kolmogorov-Arnold representation theorem, which decomposes multivariate functions into sums of univariate functions ($\Phi(\mathbf{x}) = \sum_q \phi_q (\sum_p \psi_{q,p}(x_p))$), the original KAN formulation replaces fixed activations in MLPs with learnable univariate B-splines. 

This additive structure provides a strong inductive bias for physical laws that are often expressed as truncated series expansions of non-linear terms~\citep{liu2024kan}. Notably, a shallow KAN can exactly represent any univariate polynomial, such as the cubic term in the Duffing oscillator, with sufficient spline resolution. In gray-box settings, KANs have shown potential in recovering hidden source terms in Neural ODEs \citep{koenig2024kan,ma2025integrating,daryakenari2026representation}.

Recent data-driven extensions enable symbolic residual recovery without priors, such as Structured Kolmogorov-Arnold Neural ODEs (SKANODEs) \citep{liu2025structured} for interpretable latents and symbolic equations from indirect data, and KAN-ODEs \citep{koenig2024kan} for hidden physics in ODEs/PDEs. Our work baselines the original vanilla KAN in recurrent physics-constrained settings like HRPINN. We hypothesized that replacing HRPINN's MLP residual branch with a KAN would yield superior discovery accuracy and parameter efficiency, particularly because the additive bias should naturally isolate independent physical contributions. To test this hypothesis, we selected two canonical oscillators with contrasting residual structures: the Duffing oscillator, whose residual is a univariate polynomial ($-0.3x^3$), and the Van der Pol oscillator, whose residual involves a multiplicative interaction ($(1 - x^2)v$). These oscillators were selected to represent the boundary between additive separability (Duffing) and multiplicative coupling (Van der Pol). Although KANs can theoretically represent multiplication through composition (for instance, $xy = \frac{1}{4}((x+y)^2 - (x-y)^2)  $), this requires deeper layers, raising questions about practical stability in a recurrent setting. Through carefully controlled studies, we establish a baseline for the suitability of vanilla KANs in hard-constrained recurrent architectures. While the hypothesis holds for univariate terms, our experiments reveal substantial challenges for variable interaction.



\section{Experimental Methods and Results}


We utilize the HRPINN framework, in which the residual branch $R_\theta(x, v)$ receives the normalized state $[x, v]$ and is implemented either as a standard ReLU MLP or as a B-spline KAN. The known physical dynamics and integrator structure are fixed in the recurrent update rule, while the network learns exclusively the residual manifold $R_\theta(x,v)$. This isolates discovery to the unknown terms (e.g., cubic or multiplicative damping) and enables the core integrator to be reused across related systems. Training is performed either with single-step teacher forcing or with backpropagation through time (BPTT). Performance is evaluated on held-out trajectories using test MSE and Discovery $R^2$, which measures grid-based correlation with the true residual. Rather than using KAN-specific symbolic pruning, we utilize a unified candidate-based fitting approach for both KAN and MLP. This assesses the accuracy of surface recovery independently of the symbolic extraction mechanism, providing a direct comparison of inductive bias effectiveness. 

We conducted 3 complementary studies, always with 100 seeds, to probe different aspects of KAN behavior: a configuration ablation varying grid size and sparsity to assess hyperparameter stability, a parameter-scale ablation under teacher forcing to compare efficiency, and a parameter-scale ablation with BPTT to examine the effect of recurrent unrolling. Discovery $R^2$ is calculated on a dense $100 \times 100$ grid over the phase space $x, v \in [-2.5, 2.5]$, comparing the network's predicted residual $R_\theta(x,v)$ against the analytical truth.


\begin{table}[t]
\centering
\caption{Configuration Ablation (95\% Bootstrap CI, $N=100$ seeds). $G$ and $k$ denote grid size and spline order. Config A ($G=5, k=3$) and Config F ($G=3, k=3$) represent the most stable KAN baselines. Many configurations fail severely on Van der Pol.}
\small
\begin{tabular}{lcc}
\toprule
Configuration & Duffing Discovery $R^2$ & VdP Discovery $R^2$ \\
\midrule
Config A (Baseline) & 0.835 $\pm$ 0.030 & \textbf{0.667 $\pm$ 0.037} \\
Config B (Spline-Forced) & 0.773 $\pm$ 0.097 & 0.320 $\pm$ 0.330 \\
Config C (Sparse-Low) & 0.595 $\pm$ 0.033 & -5.229 $\pm$ 5.091 \\
Config D (Sparse-High) & 0.582 $\pm$ 0.026 & -1.688 $\pm$ 1.318 \\
Config E (Aggressive-Grid) & 0.794 $\pm$ 0.067 & 0.699 $\pm$ 0.065 \\
Config F (Coarse-Grid) & \textbf{0.862 $\pm$ 0.037} & 0.639 $\pm$ 0.302 \\
Config G (Fine-Grid) & 0.745 $\pm$ 0.099 & -0.174 $\pm$ 0.691 \\
\midrule
\textbf{MLP (Small, 337 params)} & \textbf{0.957 $\pm$ 0.009} & \textbf{0.768 $\pm$ 0.015} \\
\bottomrule
\end{tabular}
\end{table}


We begin by examining the stability of KAN training across hyperparameter configurations. Table 1 reports Discovery $R^2$ for seven grid configurations over 100 random seeds. While certain coarse-grid settings achieve strong performance on the Duffing oscillator, surpassing the baseline in some cases, most configurations yield high variance or outright failure on the Van der Pol oscillator, with several producing negative $R^2$ values indicative of diverging solutions. In contrast, MLPs exhibited robust performance across standard settings. 

This extreme sensitivity underscores a practical fragility in the original KAN formulation that is absent in conventional MLPs \citep{noorizadegan2025practitioner}, justifying our choice of oscillators as representative test cases for stability. Having identified viable configurations, we next compared parameter efficiency under teacher forcing. As shown in Table 2, very small KANs ($\approx$ 120 parameters) perform competitively with similarly sized MLPs on the univariate Duffing residual and occasionally outperform larger MLPs on this task, supporting our hypothesis for separable terms. However, the picture changes for Van der Pol: even modestly wider or deeper KANs collapse, yielding near-zero or negative $R^2$, while MLPs scale gracefully with parameter count and consistently achieve higher accuracy. 

Switching to full recurrent training with BPTT partially mitigates these issues for shallow KANs (Table 2). The smallest KAN configuration achieves its highest Van der Pol score under BPTT ($R^2$ $\approx$ 0.74), suggesting that longer horizon supervision helps stabilize learning of variable interaction. Nevertheless, MLPs still dominate across nearly all scales, and deeper KANs remain catastrophically unstable, as evidenced by the high standard deviation and training failures. Qualitative examination of learned residual surfaces reinforces these quantitative findings (Figure 1). 

For the Duffing oscillator, the median small KAN accurately reproduces the characteristic cubic shape; simplified candidate fitting yielded $R^2=0.91$ for a discovered residual of $-0.234x^3$ (Ground Truth: $-0.3x^3$). While the coefficient is underestimated, the KAN captured the cubic structure in 38\% of seeds, outperforming the MLP's fitted form ($R^2=0.85$) in terms of local correlation. For Van der Pol, however, the same KAN approximates the multiplicative structure poorly, often collapsing to a roughly linear form rather than the expected parabolic modulation.


\begin{table}[ht!]
\centering
\caption{Parameter ablation — One-Step Teacher Forcing vs BPTT (Mean $\pm$ 95\% CI, N=100 seeds).}
\label{tab:ablation-compact}
\footnotesize
\setlength{\tabcolsep}{6pt}
\renewcommand{\arraystretch}{0.95}
\begin{tabular}{@{} llr  cc  cc @{}}
\toprule
\multicolumn{1}{c}{Arch.} & \multicolumn{1}{c}{Config.} & \multicolumn{1}{c}{Params}
  & \multicolumn{2}{c}{One-Step TF} & \multicolumn{2}{c}{BPTT} \\
\cmidrule(lr){4-5} \cmidrule(l){6-7}
 & & & Duffing $R^2$ & VdP $R^2$ & Duffing $R^2$ & VdP $R^2$ \\
\midrule
\multirow{4}{*}{KAN}
  & Very Small & 120  & 0.836 $\pm$ 0.032 & 0.464 $\pm$ 0.166 & \textbf{0.914 $\pm$ 0.061} & 0.743 $\pm$ 0.061 \\
  & Small      & 240  & 0.777 $\pm$ 0.079 & 0.322 $\pm$ 0.292 & 0.874 $\pm$ 0.080 & 0.785 $\pm$ 0.073 \\
  & Wide       & 480  & \textbf{0.845 $\pm$ 0.025} & 0.232 $\pm$ 0.570 & 0.468 $\pm$ 0.773 & -0.602 $\pm$ 2.842 \\
  & Deep       & 880  & -3.146 $\pm$ 7.106 & -0.303 $\pm$ 1.579 & (Unstable) & 0.754 $\pm$ 0.079 \\
\midrule
\multirow{4}{*}{MLP}
  & Tiny       & 105  & 0.914 $\pm$ 0.026 & 0.593 $\pm$ 0.048 & 0.906 $\pm$ 0.092 & 0.622 $\pm$ 0.173 \\
  & Small      & 337  & 0.957 $\pm$ 0.009 & 0.768 $\pm$ 0.015 & 0.937 $\pm$ 0.047 & 0.879 $\pm$ 0.032 \\
  & Medium     & 1185 & 0.960 $\pm$ 0.013 & 0.805 $\pm$ 0.014 & \textbf{0.951 $\pm$ 0.033} & 0.879 $\pm$ 0.019 \\
  & Large      & 4417 & \textbf{0.965 $\pm$ 0.009} & \textbf{0.843 $\pm$ 0.010} & 0.932 $\pm$ 0.063 & \textbf{0.898 $\pm$ 0.017} \\
\bottomrule
\end{tabular}

\caption*{Note: Mean $\pm$ 95\% CI, N=100 seeds. ``(Unstable)'' indicates training instability observed for deeper KANs under BPTT.}
\end{table}

\begin{figure}[ht!]
\centering
\includegraphics[width=\textwidth]{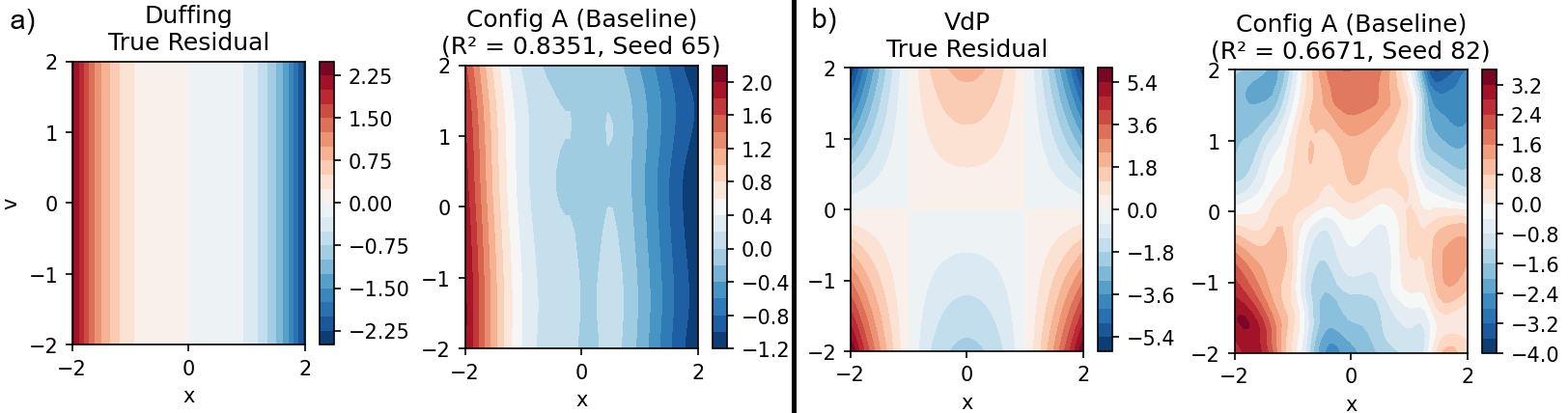}
\caption{Comparison of True Residual Surfaces vs. KAN-Recovered Manifolds (Config A). (a) Duffing: The KAN successfully identifies the univariate cubic manifold ($x^3$), demonstrating high fidelity in separable terms. (b) Van der Pol: The KAN struggles to resolve the multiplicative structure ($(1-x^2)v$), approximating the magnitude (with relevant error given the difference in the colorbars) but failing on the coupled interaction. Results show representative seeds (65 and 82) that align with the median performance reported in Table 1.}
\label{fig:heatmaps}
\end{figure}

\section{Discussion and Conclusion}

Our experiments offer a nuanced perspective on the original KAN formulation within hard-constrained physics-informed recurrent architectures. The hypothesis is validated for separable, univariate residuals: small KANs efficiently recover polynomial non-linearities with parameter counts competitive to MLPs. Yet a critical limitation emerges for multiplicative interactions. Although KANs can theoretically express multiplication through layered composition, our deeper vanilla configurations (Table 2, ``Deep" row) prove highly unstable during recurrent optimization \citep{sohail2024training}, a problem exacerbated by rapid error accumulation in the hard-constrained integration loop. MLPs, utilizing dense matrix multiplication, enforce variable interaction in the first layer ($w_ix + w_jv$). 

In contrast, vanilla KANs possess a strong additive inductive bias ($\phi(x) + \phi(v)$). To model the multiplicative Van der Pol term $(1-x^2)v$, the KAN must learn to approximate the product operator through deep composition. Our results demonstrate that even with the correct functional form provided via candidate fitting, the KAN fails to accurately recover the multiplicative manifold. This suggests that the primary bottleneck in recurrent KAN integration is the optimization stability of the composition under recurrent error accumulation, rather than the symbolic extraction process itself or a fundamental lack of expressivity, leading to the observed optimization challenges in variable interaction.

We note that subsequent KAN variants, such as deeper operator networks, hybrid formulations, or specialized scaling for oscillatory dynamics \citep{abueidda2025deepokan,zhang2025physics,mostajeran2025scaled}, may overcome these shortcomings. For instance, SKANODEs \citep{liu2025structured} and KAN-ODEs \citep{koenig2024kan} demonstrate symbolic potential in continuous Neural ODEs, motivating extensions to HRPINN for state coupling. While deeper or hybrid KAN variants may improve expressivity for multiplicative interactions, they typically incur higher computational costs and can introduce additional training instability. Nevertheless, hybrid approaches such as operator chaining (e.g., representing $1-x^2$ separately before interaction with velocity) remain promising directions for improved multiplicative stability. The present study utilizes a unified candidate-fitting benchmark to assess manifold fidelity. While this levels the comparison with MLPs, it is important to note that KANs possess a unique structural advantage: the potential for direct symbolic recovery through spline pruning, a capability entirely absent in standard MLPs. 

Our results suggest that while this latent advantage exists, its realization in recurrent physics-informed settings is currently throttled primarily by optimization instability rather than a lack of expressivity. Furthermore, while demonstrated here on ODEs, these findings on coupling stability serve as a foundational step toward recurrent architectures for PDE discretizations. Building on this initial PoC baseline for vanilla KANs in recurrent settings, future investigations should target stabilized learning of interaction terms and integrate automatic symbolic extraction to fully realize the interpretive potential of KAN-based physics-informed modeling. Additionally, future work will explore broader dynamical systems (including chaotic regimes such as the Lorenz attractor), comparisons against established symbolic methods such as SINDy, deeper diagnostics of optimization dynamics and gradient conditioning, and ultimate scaling to PDEs.



\subsubsection*{Acknowledgments}

This work was partially funded by Fundação de Apoio da UFMG (Fundep), through Linha VI – Conectividade 
Veicular, a prioritary program from Mover (Mobilidade Verde e Inovação), project AutoDL (29271.03.01/2023.04-00) and Auto5G (29271.02.01/2022.01-00).

\bibliographystyle{iclr2026_conference}
\bibliography{references}


\end{document}